\documentclass{article}
\usepackage{spconf,amsmath,epsfig}
\usepackage{url}
\usepackage{amssymb}
\usepackage[]{booktabs}
\usepackage{multirow}
\usepackage[ruled,linesnumbered]{algorithm2e}

\begin{document}\sloppy

\def\x{{\mathbf x}}
\def\L{{\cal L}}

\title{An Expressive Deep Model for Human Action Parsing from A Single Image}
%
\name{Zhujin Liang$^1$, ~~Xiaolong Wang$^1$, ~~Rui Huang$^2$, ~~Liang Lin$^{1,3,*}$\thanks{$^*$Corresponding author is Liang Lin. This work was supported by National Science Foundation of China (no. 61105014), Guangdong Science and Technology Program (no. 2012B031500006), Guangdong Natural Science Foundation
(no. S2013050014548), Special Project on Integration of Industry, Education
and Research of Guangdong Province (no. 2012B091100148), and Fundamental
Research Funds for the Central Universities (no. 13lgjc26).}}
\address{$^1$Sun Yat-Sen University, Guangzhou, China. \\ $^2$Huazhong University of Science and Techonology, Wuhan, China. \\ $^3$SYSU-CMU Shunde International Joint Research Institute, Shunde, China. \\ \{ alfredtofu, dragonwxl123 \}@gmail.com, linliang@ieee.org, ruihuang@hust.edu.cn }

\maketitle

\begin{abstract}
This paper aims at one newly raising task in vision and multimedia research: recognizing human actions from still images.  Its main challenges lie in the large variations in human poses and appearances, as well as the lack of temporal motion information.  Addressing these problems, we propose to develop an expressive deep model to naturally integrate human layout and surrounding contexts for higher level action understanding from still images.  In particular, a Deep Belief Net is trained to fuse information from different noisy sources such as body part detection and object detection.  To bridge the semantic gap, we used manually labeled data to greatly improve the effectiveness and efficiency of the pre-training and fine-tuning stages of the DBN training.  The resulting framework is shown to be robust to sometimes unreliable inputs (e.g., imprecise detections of human parts and objects), and outperforms the state-of-the-art approaches.

\end{abstract}

\section{Introduction}
\label{sec:intro}

\begin{figure}[!htpb]
\centering
\includegraphics[width=\linewidth]{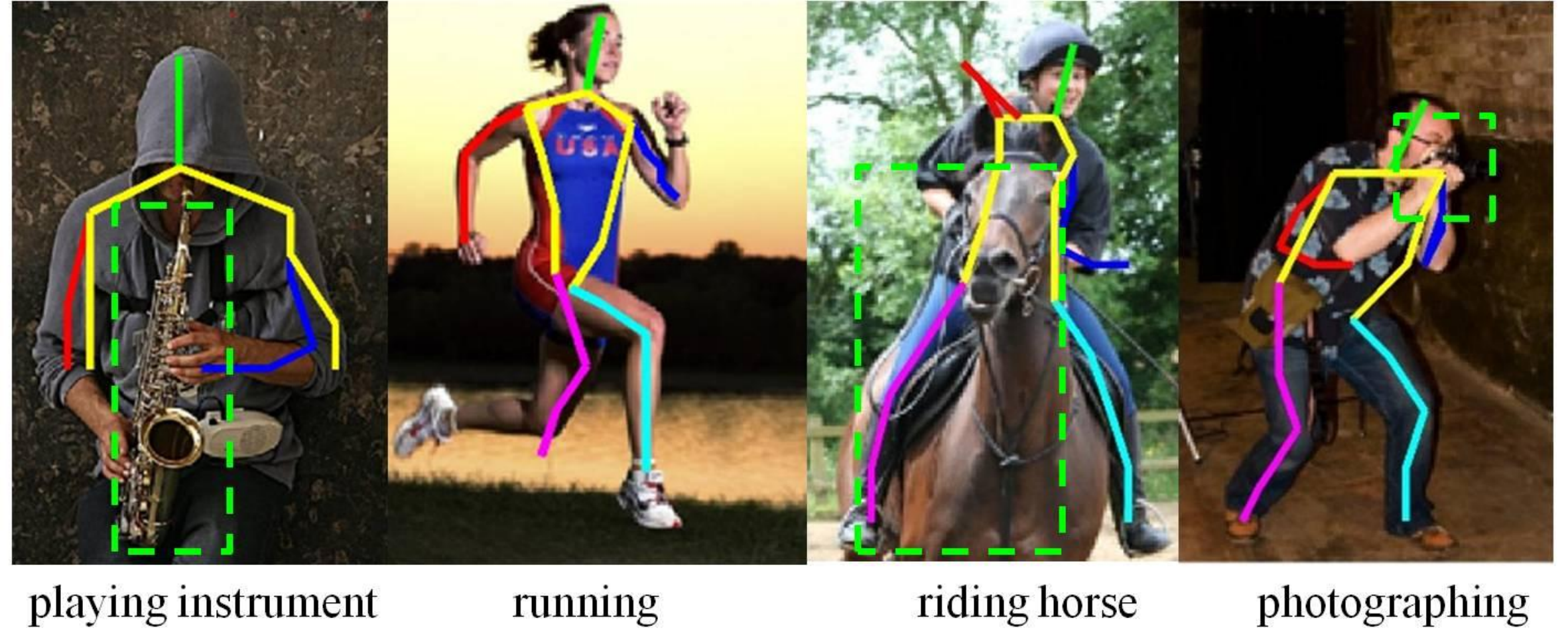}
\vspace{-8mm}
\caption{Recognizing human actions from still images, with the help of surrounding objects.}
\label{fig:parsing}
\end{figure}

This paper aims at a newly raising task in vision and multimedia research: recognizing human actions from \textit{still images} (Fig.~\ref{fig:parsing}).   Although action recognition is usually addressed in videos with motion information~\cite{Liang13,LinEvent}, more attention has been attracted to parsing human actions in still images recently for the following reasons: first, human actions represent essential content of many still images and are crucial for image understanding;  second, parsing actions in still images form the foundation of understanding complex activities;  third, not all actions contain notable dynamic information, e.g., reading a book.

The main challenges of parsing human actions in still images lie in the large human variations in poses and appearances, as well as the lack of temporal motion information.  In such a scenario, the human pose, the contextual information surrounding the human, and the interactions between the human and the surrounding contexts become crucial for understanding the action.  Traditional methods rely heavily on the accurate estimation of such information from different sources, while all these problems are themselves challenging open problems.  Moreover, the high-level human pose representations and person-object interactions are often carefully designed by hand, which is hard to generalize.

Addressing these difficulties, we propose to develop an expressive deep model to naturally integrate the information from multiple noisy, sometimes unreliable, sources such as the human layout and the surrounding objects for higher level action understanding.  In particular, a Deep Belief Net (DBN) is trained to take input of the simple features consisting of the human part detection and object detection from some off-the-shelf detectors without explicit inference of the human pose and the person-object interaction. It is worth mentioning that we applied manually labeled data of human pose and object locations during the DBN training phase, which leads to deep models learned from semantic information directly offered by human.  The trained DBN performs surprisingly well in recognizing human actions, demonstrating the capability of the deep model to learn proper feature representations of the intrinsic relationships among the simple semantic elements. 

The main contribution of the paper is 1) the proposal of developing an expressive deep model to naturally integrate human layout and surrounding contexts for higher level action understanding; 2) the use of manually labeled data to greatly improve the effectiveness and efficiency of the DBN training by feeding the deep models with semantic information provided by human; and 3) a practical state-of-the-art method for action recognition from a single image.  The resulting framework outperforms existing methods by a large margin as validated in the experiments.  It is worth noting that our approach does not rely on any specially designed problem-specific components, though we used a few off-the-shelf tools and publicly available datasets for training.  Therefore, our model can be easily generalized to solve other multimedia applications.

\section{Related Work}

A major category of methods for recognizing human actions in still images use pose and shape information of the human body in the images~\cite{XuReID13}.  In \cite{Thurau08} the authors recognized the \textit{primitive} actions using recognized \textit{pose primitives}, whereas more complex activities can be understood as a sequencing of these primitive actions.  Yang et al.\cite{Yang10} proposed to treat the pose of the person in an image as latent variables and train a system in an integrated fashion that jointly considers pose estimation and action recognition.  Rectenly, \cite{Sharma13} proposed a new Expanded Parts Model (EPM) for human analysis.  The model learns a collection of discriminative templates which can appear at specific scale-space positions.   A more recent work by Khan et al.~\cite{Khan13} combines color and shape information for action recognition.  They perform a comprehensive evaluation of color descriptors and fusion approaches and suggests that incorporating color
information considerably improves recognition performance, a descriptor based on color names
outperforms pure color descriptors, late fusion of color and shape information outperforms other approaches, and different fusion approaches result in complementary information which should be combined.  These two works~\cite{Sharma13, Khan13} have reported the current state-of-the-art performances in this task.  The methods in this category usually rely heavily on pose and shape representation and estimation, which are often severely affected by illumination, occlusions, viewing angle, etc.

Another category of methods discriminate different actions in still images using contextual information, especially the objects surrounding the human subject and the person-object interactions.  Gupta et al.~\cite{Gupta09} present a Bayesian approach which goes beyond static shape/appearance feature matching and motion analysis used in traditional object and action recognition, and applies spatial and functional constraints on each of the perceptual elements for coherent semantic interpretation.  This approach works even when the appearances and motion information are not discriminative enough.  Desai et al.~\cite{Desai10} advocate an approach to activity recognition based on modeling contextual interactions between postured human bodies and nearby objects.  Similarly, \cite{Yao12} proposes a mutual context model to jointly model objects and human poses in human-object interaction activities. In this approach, object detection provides a strong prior for better human pose estimation, while human pose estimation improves the accuracy of detecting the objects that interact with the human.  Building on locally order-less spatial pyramid bag-of-features model, Delaitre et al. investigated a discriminatively trained model of person-object interactions for recognizing common human actions in still images~\cite{Delaitre10}.  They replace the standard quantized local HOG/SIFT features with stronger discriminatively trained body part and object detectors, introduce new person-object interaction features based on spatial co-occurrences of individual body parts and objects, and address the combinatorial problem of a large number of possible interaction pairs and propose a discriminative selection procedure using a linear support vector machine (SVM) with a sparsity inducing regularizer.  These approaches also rely on accurate pose estimation, and in addition, object detection.  Moreover, the person-object interactions are often carefully designed by hand, which is hard to generalize.  Although the most recently reported state-of-the-art resutls~\cite{Sharma13,Khan13} belong to the first category of methods, we believe the contextual objects do provide valuable information in parsing human actions and thus included them in our framework.

Recently, deep models~\cite{Hinton06,Bengio09} have shown excellent capability in learning image representations without using hand-crafted features.  Deep learning has been successfully applied in many vision problems such as image classification~\cite{Krizhevsky12} and action recognition in videos~\cite{Ji10}.  These methods directly applied deep models to learn feature representations from raw data.  Recently, Ouyang et al.~\cite{Ouyang12} proposed a deep model that takes the human part detection results as input and learns the relationships between human parts to handle the occlusion problems in pedestrian detection.  It motivates us to learn the mutual context between objects and human parts with deep models, which has not been studied before.  Deep belief nets~\cite{Hinton06} are probabilistic generative models that are composed of multiple layers of stochastic, latent variables.  The latent variables typically have binary values and are often called hidden units or feature detectors.  The top two layers have undirected, symmetric connections between them and form an associative memory.  The lower layers receive top-down, directed connections from the layer above.  The states of the units in the lowest layer represent a data vector.  DBNs were first developed for binary data using a Restricted Boltzmann Machine (RBM) as the basic module for learning each layer.  The hidden and visible biases and the matrix of weights connecting the visible and hidden units are easy to train using contrastive divergence learning which is a crude but efficient approximation to maximum likelihood learning\cite{Hinton06}.  

In this paper, we propose to train a DBN, as shown in Fig.~\ref{fig:DBN} to alleviate the dependence of action recognition on accurate pose estimation and to more naturally incorporate contextual object information.  In particular, the DBN fuses the information of the human layout, the contextual objects surrounding the human, and the person-object interactions in still images to recognize the human actions.  Note that these pieces of information are often noisy and sometimes unreliable.  In the experiments we show that the trained DBN outperforms the state-of-the-art approaches using automatically detected human parts and surrounding objects.

\section{Proposed method}

\begin{figure}[htpb]
\centering
\includegraphics[width=\linewidth]{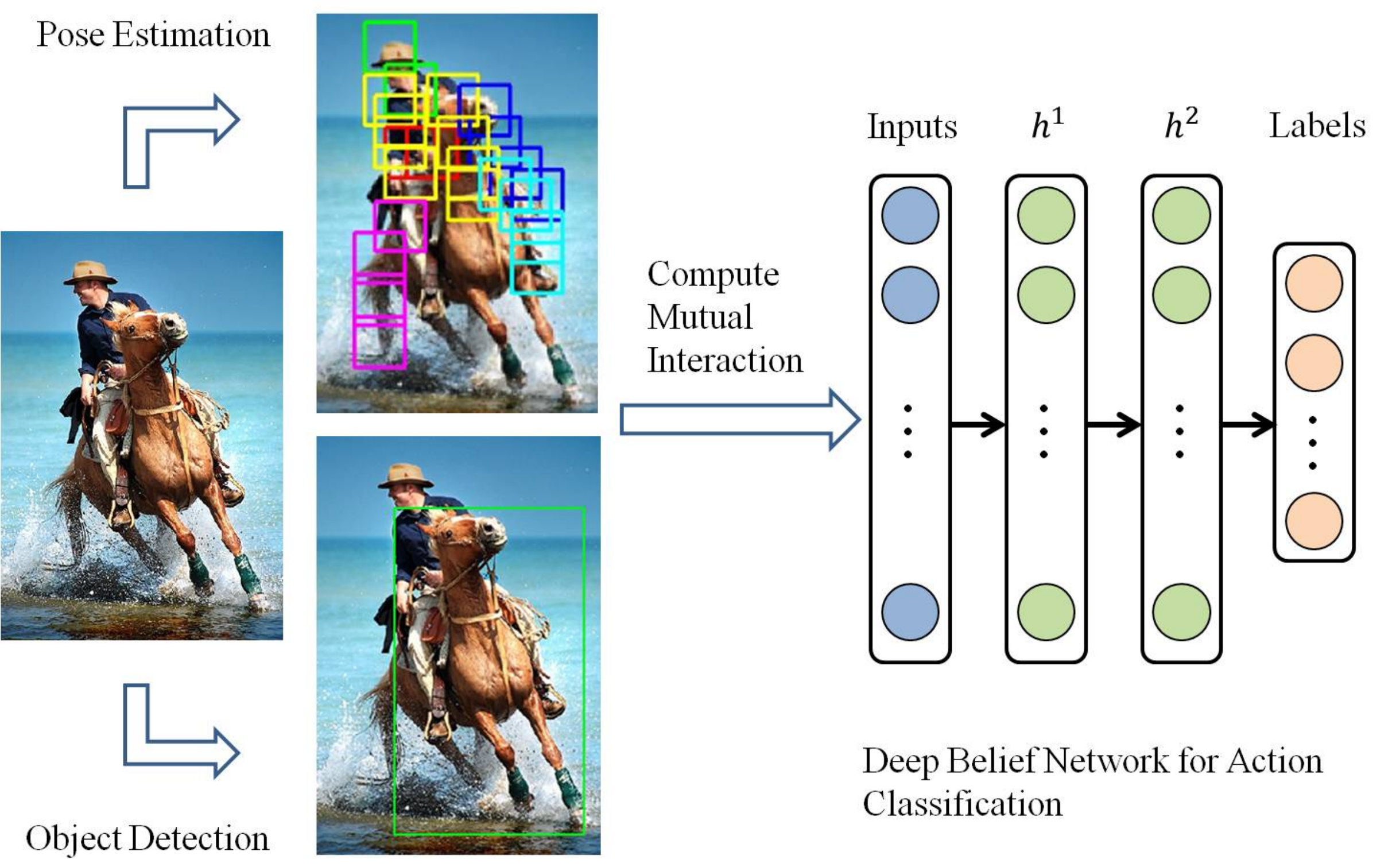}
\vspace{-8mm}
\caption{The proposed framework. Given a image, we first performed human part detection and object detection on it.  We then extracted the features of mutual interactions between objects and human parts. Given the mutual contextual features as input, we applied the Deep Belief Net for action recognition.}
\label{fig:framework}
\end{figure}

The proposed method is shown in Fig.~\ref{fig:framework}.  Note that the training phase and the testing phase follow the similar procedure, except that during the training stage, the body part detectors and object detectors are first trained using some off-the-shelf tools and publicly available datasets, the detection results and some manually labeled detections are used as the input to the DBN, and the DBN parameters are then learned.  In the testing stage, the human parts and objects are detected automatically by the trained detectors and used as input by the DBN to predict the action type.  We explain the procedure in details in the rest of this section.

\subsection{Body Part Detection and Pose Estimation}

\begin{figure}[htpb]
\centering
\includegraphics[width=\linewidth]{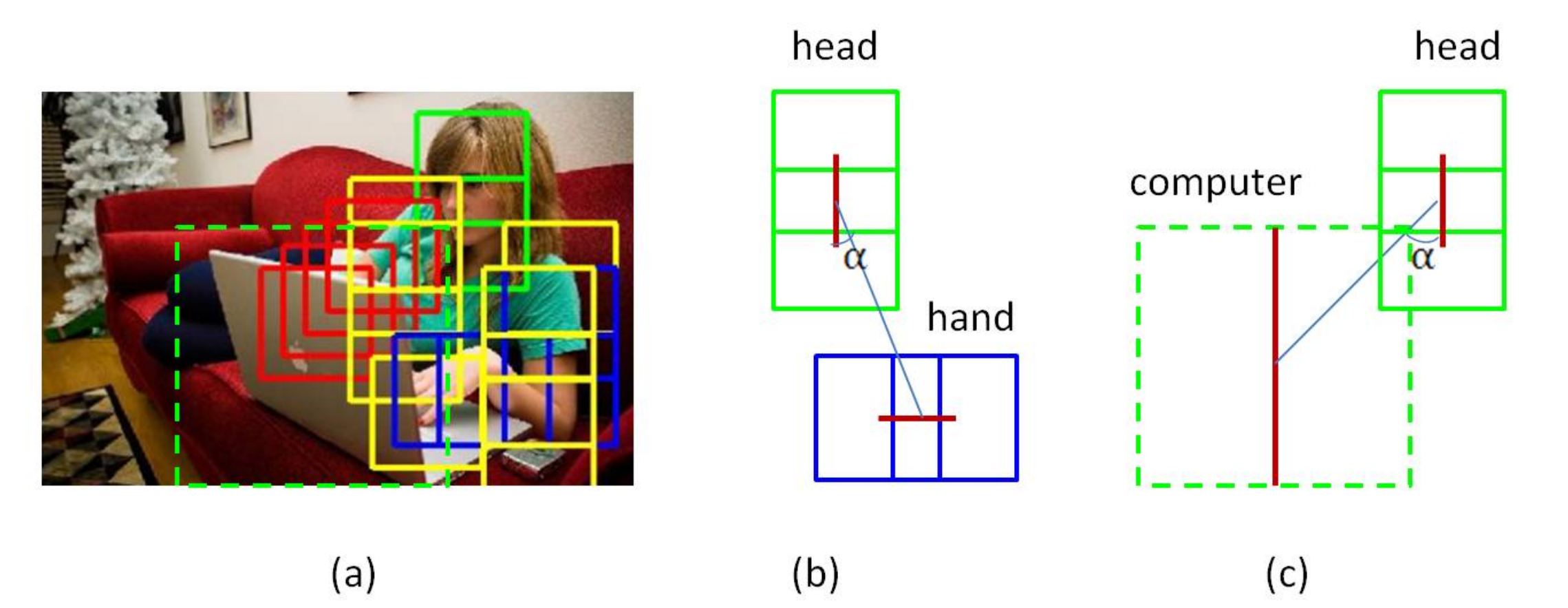}
\vspace{-10mm}
\caption{Human part and contextual object representation: (a)shows a pose estimation and object detection result performed on the image; (b)demonstrates the spatial relationships between human head and hand. (c)demonstrates the spatial relationships between human head and computer.  The red line across two rectangles is a central line for a human part, and it is not necessarily vertical or horizontal. The central line of an object is the red line across the detection window. $\alpha$ is the angle between the central line of the head and the line connecting the head to the body part or the object.}
\label{fig:pose}
\end{figure}

\begin{figure}[htpb]
\centering
\includegraphics[width=\linewidth]{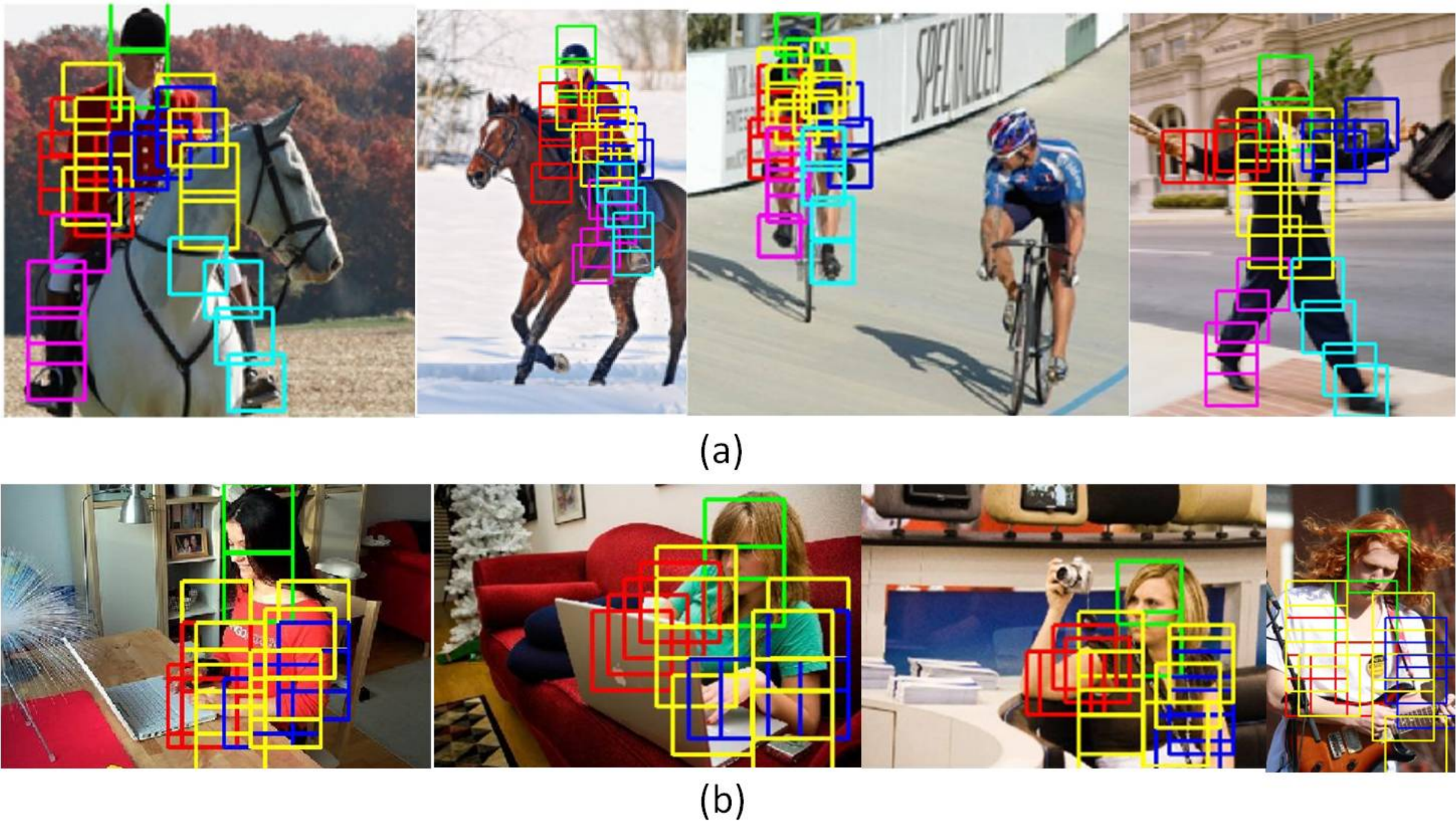}
\vspace{-10mm}
\caption{Human part detection.(a) shows the results performed by full body part detection (with 10 parts); (b) shows the results performed by upper body part detection (with 6 parts).}
\label{fig:partdet}
\end{figure}

For each input image, we first detect 10 parts of a human body using the method described in~\cite{XuReID13} (as shown in Fig.~\ref{fig:partdet}(a)).  Pose representation and estimation is a challenging open problem.  In our framework, since we rely mostly on the capability of the deep model to learn proper representations from the high dimensional data, the human pose is loosely represented as a star model, where the head is the centroid and the normalized relative locations of all the other parts are computed as the features.  As shown in Fig.~\ref{fig:pose}(b), a human part is defined by two small part detectors.  The central line of the human part is the line between the centroids of the two part detectors(the red line).  We first normalize the image size to fix the length of the head to 50 pixels.  For any other body part, a 6 dimensional feature vector is computed as

\begin{equation}
[isExist,x_1,y_1,x_2,y_2,\alpha],
\end{equation}

\noindent where $isExist$ indicates whether the interaction between the part and the head exists, $x_1$, $y_1$, $x_2$, $y_2$ are the coordinates (relative to the center of the head) of the central line of the part, and $\alpha$ the angle between the central line of the head and the line connecting the head to the part.

In our implementation, we also applied the upper body parts detector introduced in~\cite{XuReID13}, which include 6 human parts out of the 10 full body human parts(as illustrated in Fig.~\ref{fig:partdet}(b)).  Given a image, we performed the full body as well as upper body pose estimation, and chose the detection result with the higher pose estimation score.  We use the variable $isExist = 0$ to indicate the missing of 4 body parts in the upper body pose estimation result. Note that we do not need to further estimate the human poses.

\subsection{Object Detection and Person-Object Relationship}

\begin{figure}[phtb]
\centering
\includegraphics[width=\linewidth]{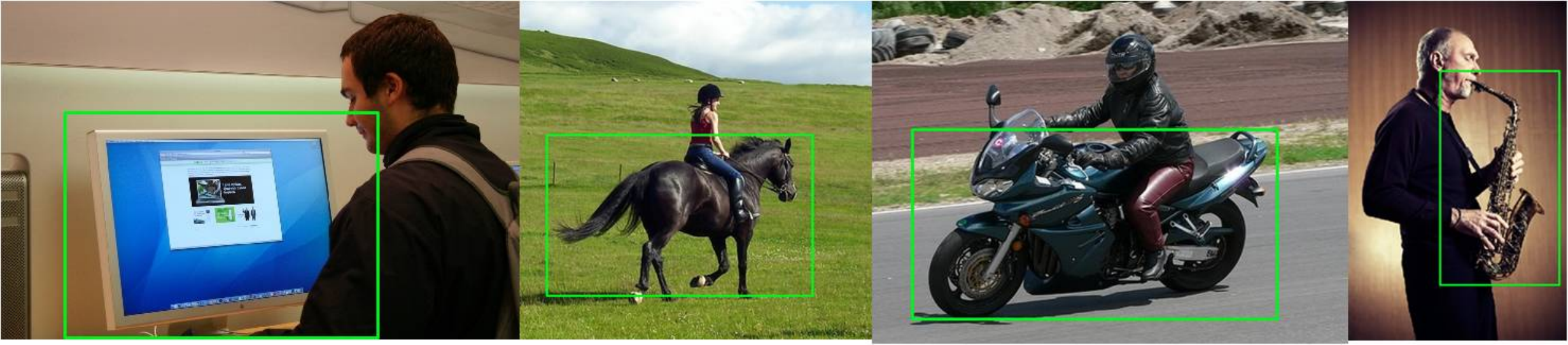}
\vspace{-8mm}
\caption{Object detection results performed by Deformable Part Models.}
\label{fig:objectdet}
\end{figure}

As mentioned previously, the contextual information is important for recognizing human actions~\cite{LinGrammar}, especially the person-object interactions.  Therefore, we train Deformable Part Models (DPMs)~\cite{Felzenszwalb10} to detect objects surrounding the human.  DPMs are the combination of 1) strong low-level features based on histograms of oriented gradients (HOG); 2) efficient matching algorithms for deformable part-based models (pictorial structures); 3) discriminative learning with latent variables (latent SVM), resulting in efficient object detectors that achieve state of the art results in practice.  In our experiments, we detect 5 types of objects (i.e., ``bike'', ``camera'', ``computer'', ``horse'', ``instrument'') using trained DPM object detectors, Fig.~\ref{fig:objectdet} shows some examples.  We again describe the person-object interactions as the relative locations of the object, and the features are similar to the ones describing the human pose.  For each object, a feature vector is computed as: $[isExist,x_1,y_1,x_2,y_2,\alpha]$, where $isExist$ indicates whether the person-object interaction between the object and the human exists, $x_1$, $y_1$, $x_2$, $y_2$ are the coordinates (relative to the center of the head) of the central line of the object, and $\alpha$ the angle between the central line of the head and the line connecting the head to the object (as illustrated in Fig.~\ref{fig:pose}(c)).  This is a very simple yet crude representation of person-object interactions.  However it works quite well with the deep model.

\subsection{Deep Belief Nets}

From the body part detection and object detection, we obtain in total 10 body parts and 5 objects.  As we model human pose as pairwise relationships between the head and the other body parts, and person-object interactions pairwise relationships between the head and the objects, there are in total $15\times6=90$ dimensions for all the features.  Each dimension of the feature is first normalized to the scale of $0 ~ 1$ with sigmoid function.

\begin{figure}[phtb]
\centering
\includegraphics[width=\linewidth]{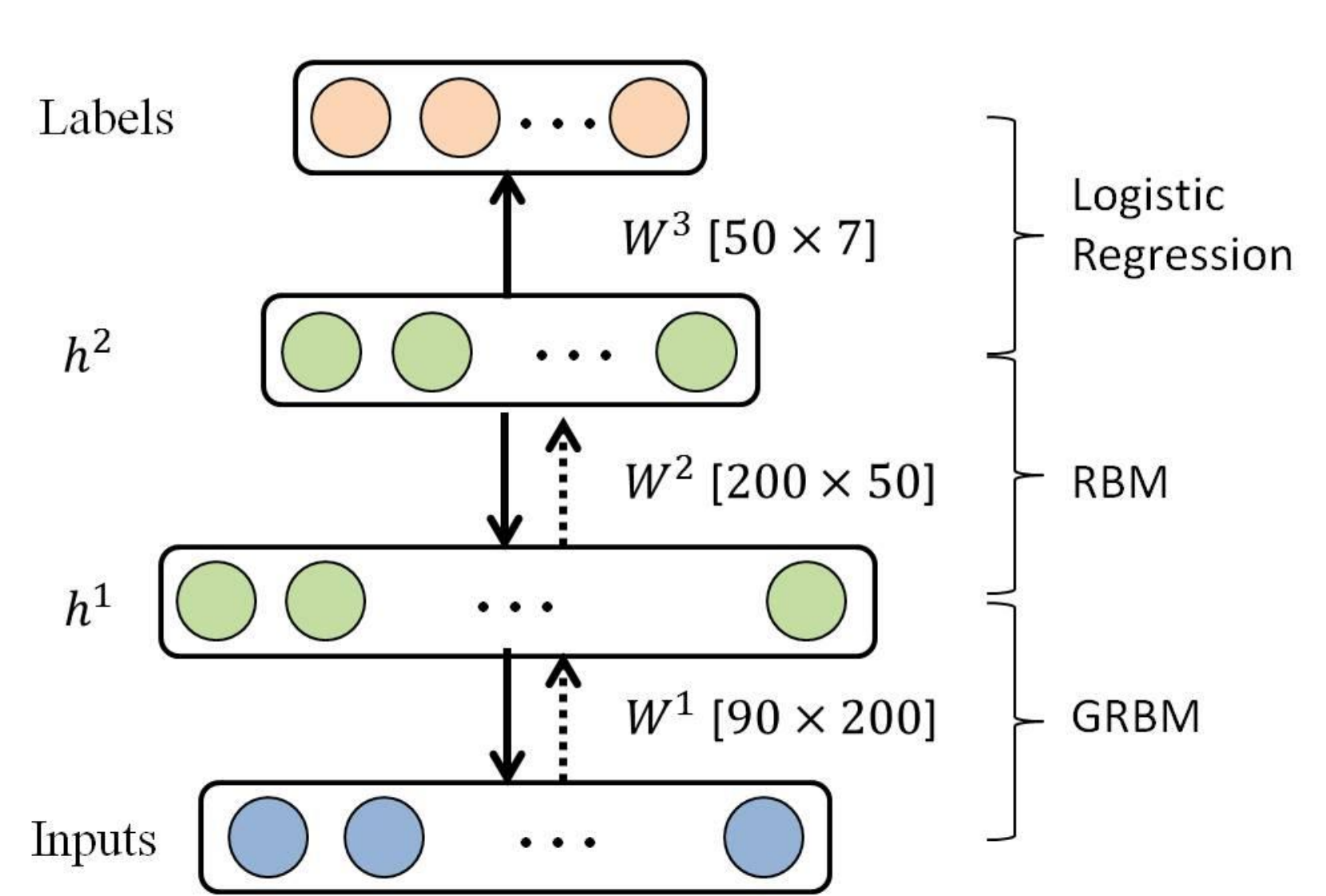}
\vspace{-8mm}
\caption{The Deep Belief Net Model. It is a four layer model with 90 neurons as input, 200 hidden variables in the second layer, 50 hidden variables in the third layer and 7 outputs as labels in the top layer. The DBN is pre-trained by stacking the Gaussian RBM and ordinary binary RBM. Fine-tunning is performed with logistic regression to optimize the all the parameters for classification.}
\label{fig:DBN}
\end{figure}

In the training phase, the feature vectors calculated from all the training images are input to the DBN, as shown in Fig.~\ref{fig:DBN}.  The DBN includes 4 layers with 90 dimensions of input in the first layer, 200 hidden variables in the second layer, 50 hidden variables in the third layer and 7 variables as output labels in the top layer.  We follow the method introduced in \cite{Hinton06} to train our DBN with layer-wised RBM pre-training and logistic regression fine-tunning.  Given the real valued data as input, we employed the Gaussian RBM(GRBM) to model the parameters between the first layer and second layer during pre-training.  The energy function of the Gaussian RBM is defined as,

\begin{equation}\label{eq:GRBM}
E(\textbf{v}, \textbf{h}) = \frac{1}{2 {\sigma}^{2}} \sum_i (v_i-c_i)^{2} - \frac{1}{\sigma}  \sum_{i,j} v_i W_{ij} h_j - \sum_{j} b_j h_j
\end{equation}

\noindent where $W$ is the model parameters, $\textbf{v}$ is the visible layer vector, $\textbf{h}$ is the hidden layer vector.  $c_i$ and $b_j$ are the biases for the visible and hidden neurons.  Given the output of binary valued data from the second layer, we built up a regular RBM upon it, whose energy function is,

\begin{equation}
E(\textbf{v}, \textbf{h}) = -\sum_{i,j} v_i W_{ij} h_j - \sum_{j} b_j h_j - \sum_{i} c_i v_i
\end{equation}

\noindent where the notations are defined similar to the Gaussian RBM.

An critical difference between our model and other works is that we used manually labeled data(human part and object locations) to greatly improve the effectiveness and efficiency of the pre-training and fine-tuning during the training phase, though the testing is performed with automatic body part detectors and object detectors trained using off-the-shelf tools and some publicly available datasets.  The experimental results show that the performance increase is obvious.

\section{Experimental Results}
\label{sec:exp}

We first introduce the datasets and the parameter settings of our implementation, and then show the experimental results.

\subsection{Dataset and settings}

\begin{figure}[htpb]
\centering
\includegraphics[width=\linewidth]{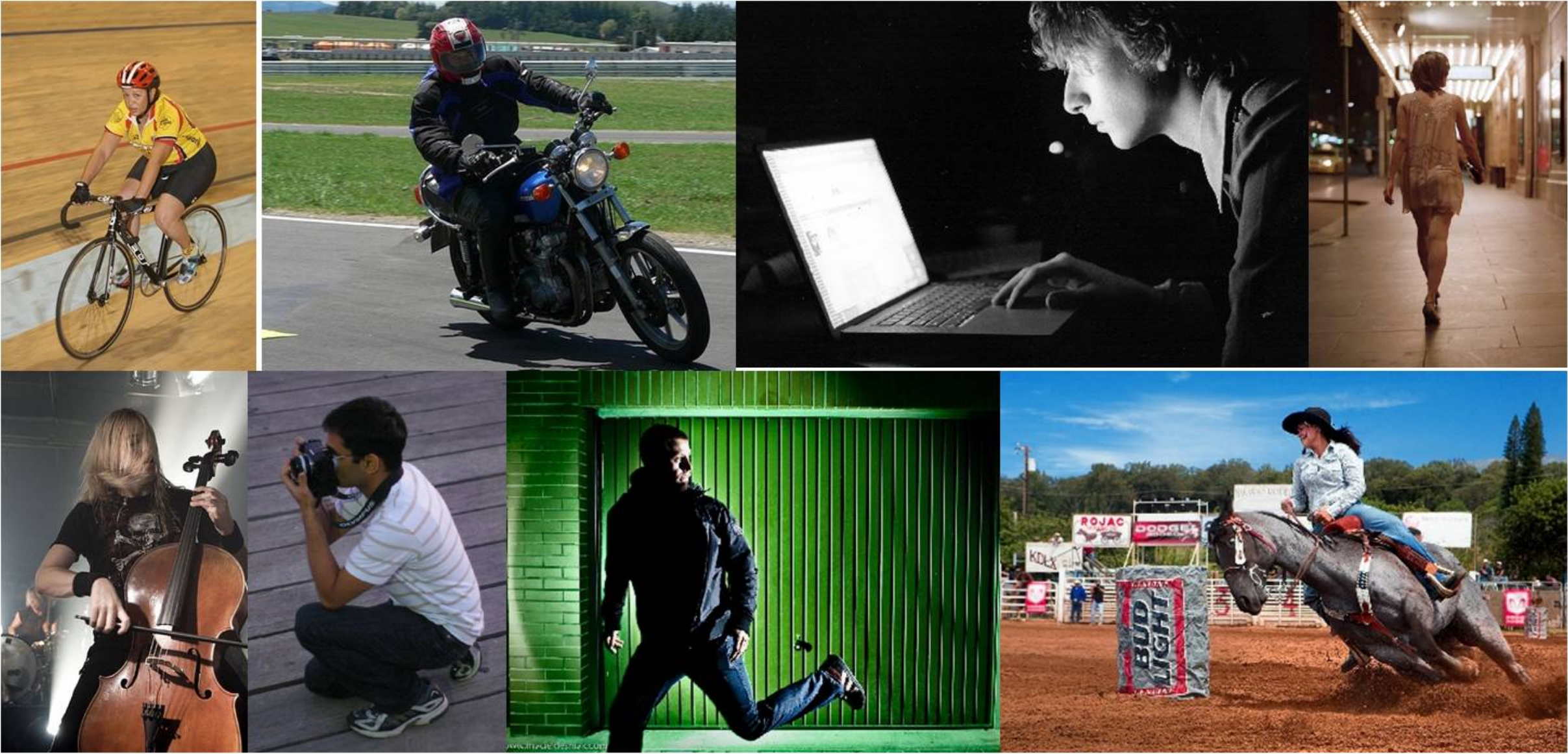}
\vspace{-8mm}
\caption{Samples of images in the Willow Action dataset.}
\label{fig:willow}
\end{figure}

We investigated the performance of the proposed method on the publicly available Willow-Actions Dataset\cite{Delaitre10}.  The Willow-Actions Dataset (Fig.~\ref{fig:willow}) is a dataset for human action classification in still images.  Action classes include ``interacting with computer'', ``photographing'', ``playing instrument'', ``riding bike'', ``riding horse'', ``running'', ``walking''.  Following the evaluation protocol in~\cite{Delaitre10}, we used 427 images for training and 484 images for testing.

\emph{Training Settings.} As mentioned before, we trained DPMs for the 5 classes.  Since ``bike'' and ``instrument'' have many subclasses, we train these two detectors using images from Imagenet\footnote{\url{http://www.image-net.org/}}, including ``bicycle'' and ``motorcycle'' for ``bike'', and ``saxophone'', ``violin'', ``piano'', ``guitar'', ``flute'', ``cello'' for ``instrument''.  The body part detectors are trained on a large dataset: Leeds Sport Pose Dataset\footnote{\url{http://www.comp.leeds.ac.uk/mat4saj/lsp.html}}.

Before the DBN training, we first flipped every image horizontally, and the human parts and objects are localized manually for the DBN training.  We also perturb the human part and object locations with a random distance of $\varepsilon$ pixels($-10 \leqq \varepsilon \leqq 10$ ) in each training image.  We repeated this procedure 10 times and generated 10 training samples from each image.  Thus we have in total $427 \times 2 \times 10 = 8540$ training samples for the DBN.  This way, we relieve the over-fitting problem and can handle the variances given by the unstable detection results during the testing.

The parameters for DBN are set as: the pre-training learning rate is 0.01, pre-training iteration number 100, fine-tuning learning rate 0.1, fine-tuning iteration number 1000.

\emph{Testing Setting.} During testing, the human pose and objects are localized by automatic body part detectors and object detectors.  A parameter $\sigma_k( 1 \leqq k \leqq 15)$ is defined for each human part and object part. For each part, if its detection score is larger than $\sigma$, we set $isExist = 1$, otherwise $isExist = 0$.  $\sigma$ is estimated by cross validation.

\subsection{Results and comparisons}

We first show the mean Average Precision (mAP) of the proposed method in recognizing the 7 classes of actions in still images and the comparisons to the state-of-the-art methods (Table~\ref{tab:Acc}).  Our DBN method reached the mAP of $80.41\%$, which is significantly (about $10\%$) higher than the state-of-the-art result of $70.1\%$~\cite{Khan13}.  Meanwhile, we also switch the DBN model with SVM (a ``shallow'' model~\cite{Bengio09}) by keeping the same input features.  Table~\ref{tab:Acc} shows that our method works much better with the deep model than with the shallow model.  

\begin{table}[htpb]
\small
  \centering

  \caption{mAP comparisons to the state of the art}

    \begin{tabular}{cc}
    \toprule
    Method  & mAP\\
    \midrule
    Our DBN   & \textbf{80.41} \\
    Our SVM   & 77.82 \\
    Khan et al.~\cite{Khan13} & 70.1 \\
    Sharma et al.~\cite{Sharma13} & 67.6 \\
    Delaitre et al.~\cite{Delaitre10} & 59.6\\

    \bottomrule
    \end{tabular}%
  \label{tab:Acc}%
\end{table}%

To show the benefits of incorporating contextual object information, we show in Table~\ref{tab:mAP} the mAP of the proposed method (DBN) and SVM, with or without using object informaiton, respectively, for the 7 different actions.  It is obvious that the contextual objects bring valuable information into the recognition of human actions.

\begin{table}[htpb]

  \centering

  \caption{mAP comparisons for different classes}

    \begin{tabular}{ccccc}

    \toprule

    Class Name & SVMw/o & SVM   & DBNw/o & DBN \\

    \midrule

    Int. Computer & 14.32 & 83.71 & 40.89 & \textbf{86.56} \\

    Photographing & 32.8  & 89.47 & 28.27 & \textbf{90.5} \\

    PlayingMusic & 39.16 & \textbf{95.95} & 40.69 & 89.91 \\

    RidingBike & 63.06 & 97.35 & 32.27 & \textbf{98.17} \\

    RidingHorse & 43.08 & 92.7  & 37.47 & \textbf{92.72} \\

    Running & 39.28 & 37.9  & 26.23 & \textbf{46.16} \\

    Walking & 61.25 & 48.79 & 40.45 & \textbf{58.88} \\

    mAP   & 41.85 & 77.98 & 35.18 & \textbf{80.41} \\

    \bottomrule

    \end{tabular}%

  \label{tab:mAP}%

\end{table}%

\begin{figure}[htpb]
\centering
\includegraphics[width=\linewidth]{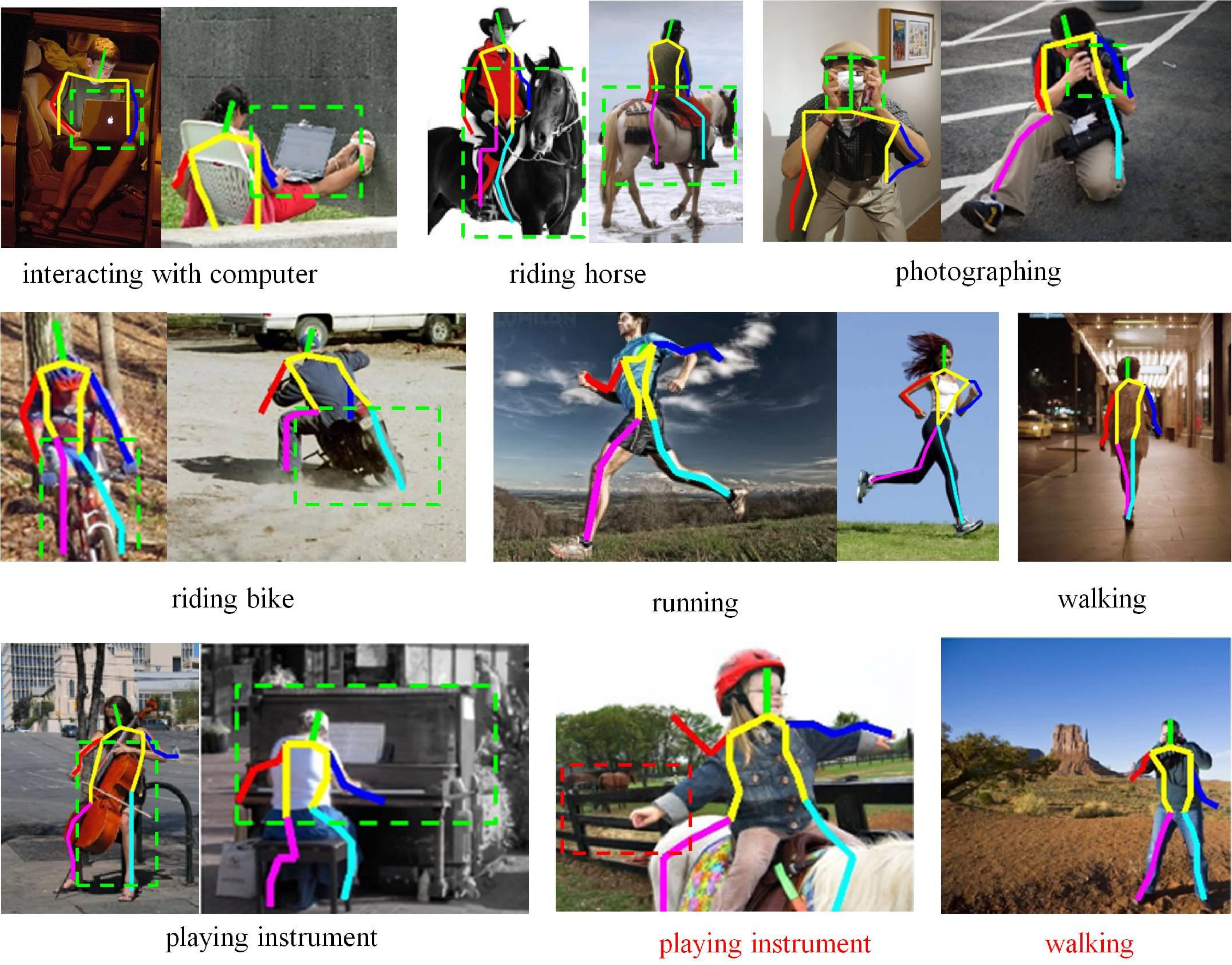}
\vspace{-9mm}
\caption{Sample results of parsing actions from still images using our framework.}
\label{fig:result}
\end{figure}

We visualize part of our results as Fig.~\ref{fig:result}) including 7 categories of action classes and two failing cases in the last line. It is shown that our model is robust even though the detections of pose and object are not reliable.  The error of classification is mainly because of the wrong object localization and the estimated pose is similar to the one in other classes.

\section{Conclusions}

We investigated the use of deep learning techniques in the task of recognizing human actions from still images.  An expressive deep model was developed to naturally integrate different sources of information, including human layout and surrounding contexts, for human action parsing in a single image.  In particular, a Deep Belief Net is trained and manually labeled training data greatly improved the effectiveness and efficiency of the pre-training and fine-tuning stages of the DBN training phase.  

\bibliographystyle{IEEEbib}
\bibliography{action}

\end{document}